\pdfoutput=1

\documentclass[11pt]{article}

\usepackage{naacl2021}

\usepackage{amsmath}
\usepackage{amssymb}
\usepackage{algorithm}
\usepackage{algorithmic}
\usepackage{booktabs}
\usepackage{caption}
\usepackage{CJKutf8}
\usepackage{graphicx}
\usepackage{multirow}
\usepackage{url}
\usepackage{subfigure}

\usepackage{times}
\usepackage{latexsym}

\usepackage[T1]{fontenc}

\usepackage[utf8]{inputenc}

\usepackage{microtype}

\title{Pre-training with Meta Learning for Chinese Word Segmentation}

\author{
  Zhen Ke\textsuperscript{1}, Liang Shi\textsuperscript{1}, Songtao Sun\textsuperscript{1}, Erli Meng\textsuperscript{1}, Bin Wang\textsuperscript{1}, Xipeng Qiu\textsuperscript{2} \\
  Xiaomi AI Lab, Xiaomi Inc., Beijing, China\textsuperscript{1} \\
   Shanghai Key Laboratory of Intelligent Information Processing, Fudan University\textsuperscript{2} \\
  \texttt{\{kezhen,shiliang1,sunsongtao,mengerli,wangbin11\}@xiaomi.com} \\
  \texttt{\{xpqiu\}@fudan.edu.cn} \\
}

\begin{document}
\begin{CJK}{UTF8}{gbsn}
\maketitle
\begin{abstract}
    Recent researches show that pre-trained models (PTMs) are beneficial to Chinese Word Segmentation (CWS).
    However, PTMs used in previous works usually adopt language modeling as pre-training tasks, lacking task-specific prior segmentation knowledge and ignoring the discrepancy between pre-training tasks and downstream CWS tasks.
    In this paper, we propose a CWS-specific pre-trained model \textsc{Meta}\textsc{Seg}, which employs a unified architecture and incorporates meta learning algorithm into a multi-criteria pre-training task.
    Empirical results show that \textsc{Meta}\textsc{Seg} could utilize common prior segmentation knowledge from different existing criteria and alleviate the discrepancy between pre-trained models and downstream CWS tasks.
    Besides, \textsc{Meta}\textsc{Seg} can achieve new state-of-the-art performance on twelve widely-used CWS datasets and significantly improve model performance in low-resource settings.
\end{abstract}

\section{Introduction}
\label{sec:intro}

Chinese Word Segmentation (CWS) is a fundamental task for Chinese natural language processing (NLP), which aims at identifying word boundaries in a sentence composed of continuous Chinese characters.
It provides a basic component for other NLP tasks like named entity recognition~\cite{li-2020-flat}, dependency parsing~\cite{yan-2020-graph}, and semantic role labeling~\cite{xia-2019-syntax}, etc.

Generally, most previous studies model the CWS task as a character-based sequence labeling task \cite{xue-2003-chinese,zheng-2013-deep,chen-2015-long,ma-2018-state,qiu-2019-multi}.
Recently, pre-trained models (PTMs) such as BERT \cite{devlin-2019-bert} have been introduced into CWS tasks, which could provide prior semantic knowledge and boost the performance of CWS systems.
\newcite{yang-2019-bert} directly fine-tunes BERT on several CWS benchmark datasets.
\newcite{huang-2019-toward} fine-tunes BERT in a multi-criteria learning framework, where each criterion shares a common BERT-based feature extraction layer and owns a private projection layer.
\newcite{meng-2019-glyce} combines Chinese character glyph features with pre-trained BERT representations.
\newcite{tian-2020-improving} proposes a neural CWS framework WM\textsc{Seg}, which utilizes memory networks to incorporate wordhood information into the pre-trained model ZEN~\cite{diao-2019-zen}.

PTMs have been proved quite effective by fine-tuning on downstream CWS tasks.
However, PTMs used in previous works usually adopt language modeling as pre-training tasks.
Thus, they usually lack task-specific prior knowledge for CWS and ignore the discrepancy between pre-training tasks and downstream CWS tasks.

\begin{table}[t]
\centering
\begin{tabular}{|c|c|c|c|c|c|}
\hline
Criteria & Li & Na & entered & \multicolumn{2}{c|}{the semi-final} \\ \hline
CTB6 & \multicolumn{2}{c|}{李娜} & 进入 & \multicolumn{2}{c|}{半决赛} \\ \hline
PKU & 李 & 娜 & 进入 & 半 & 决赛 \\ \hline
MSRA & \multicolumn{2}{c|}{李娜} & 进入 & 半 & 决赛 \\ \hline
\end{tabular}
\caption{An example of CWS on different criteria.}
\label{tbl:example}
\end{table}

To deal with aforementioned problems of PTMs, we consider introducing a CWS-specific pre-trained model based on existing CWS corpora, to leverage the prior segmentation knowledge.
However, there are multiple inconsistent segmentation criteria for CWS, where each criterion represents a unique style of segmenting Chinese sentence into words, as shown in Table~\ref{tbl:example}.
Meanwhile, we can easily observe that different segmentation criteria could share a large proportion of word boundaries between them, such as the boundaries between word units ``李娜(Li Na)'', ``进入(entered)'' and ``半决赛(the semi-final)'', which are the same for all segmentation criteria.
It shows that the common prior segmentation knowledge is shared by different criteria.

In this paper, we propose a CWS-specific pre-trained model \textsc{Meta}\textsc{Seg}.
To leverage shared segmentation knowledge of different criteria, \textsc{Meta}\textsc{Seg} utilizes a unified architecture and introduces a multi-criteria pre-training task.
Moreover, to alleviate the discrepancy between pre-trained models and downstream unseen criteria, meta learning algorithm~\cite{finn-2017-maml} is incorporated into the multi-criteria pre-training task of \textsc{Meta}\textsc{Seg}.

Experiments show that \textsc{Meta}\textsc{Seg} could outperform previous works significantly, and achieve new state-of-the-art results on twelve CWS datasets.
Further experiments show that  \textsc{Meta}\textsc{Seg} has better generalization performance on downstream unseen CWS tasks in low-resource settings, and improve Out-Of-Vocabulary (OOV) recalls.
To the best of our knowledge, \textsc{Meta}\textsc{Seg} is the first task-specific pre-trained model especially designed for CWS.

\section{Related Work}
\label{sec:related}

Recently, PTMs have been used for CWS and achieve good performance~\cite{devlin-2019-bert}.
These PTMs usually exploit fine-tuning as the main way of transferring prior knowledge to downstream CWS tasks.
Specifically, some methods directly fine-tune PTMs on CWS tasks~\cite{yang-2019-bert}, while others fine-tune them in a multi-task framework~\cite{huang-2019-toward}.
Besides, other features are also incorporated into PTMs and fine-tuned jointly, including Chinese glyph features~\cite{meng-2019-glyce}, wordhood features~\cite{tian-2020-improving}, and so on.
Although PTMs promote CWS systems significantly, their pre-training tasks like language modeling still have a wide discrepancy with downstream CWS tasks and lack CWS-specific prior knowledge.

Task-specific pre-trained models are lately studied to introduce task-specific prior knowledge into multiple NLP tasks.
Specifically designed pre-training tasks are introduced to obtain the task-specific pre-trained models, and then these models are fine-tune on corresponding downstream NLP tasks, such as named entity recognition~\cite{xue-2020-coarsetofine}, sentiment analysis~\cite{ke-2020-sentilare} and text summarization~\cite{zhang-2020-pegasus}.
In this paper, we propose a CWS-specific pre-trained model \textsc{Meta}\textsc{Seg}.

\section{Approach}
\label{sec:approach}

As other task-specific pre-trained models~\cite{ke-2020-sentilare}, the pipeline of \textsc{Meta}\textsc{Seg} is divided into two phases: pre-training phase and fine-tuning phase.
In pre-training phase, we design a unified architecture and incorporate meta learning algorithm into a multi-criteria pre-training task, to obtain the CWS-specific pre-trained model which has less discrepancy with downstream CWS tasks.
In fine-tuning phase, we fine-tune the pre-trained model on downstream CWS tasks, to leverage the prior knowledge learned in pre-training phase.

In this section, we will describe \textsc{Meta}\textsc{Seg} in three parts.
First, we introduce the Transformer-based unified architecture.
Second, we elaborate on the multi-criteria pre-training task with meta learning algorithm.
Finally, we give a brief description of the downstream fine-tuning phase.

\subsection{The Unified Architecture}
\label{subsec:unified}


In traditional CWS systems~\cite{chen-2015-long,ma-2018-state}, CWS model usually adopts a separate architecture for each segmentation criterion.
An instance of the CWS model is created for each criterion and trained on the corresponding dataset independently.
Thus, a model instance can only serve one criterion, without sharing any segmentation knowledge with other different criteria.

\begin{figure*}[t]
    \centering
    \includegraphics[width=0.99\textwidth]{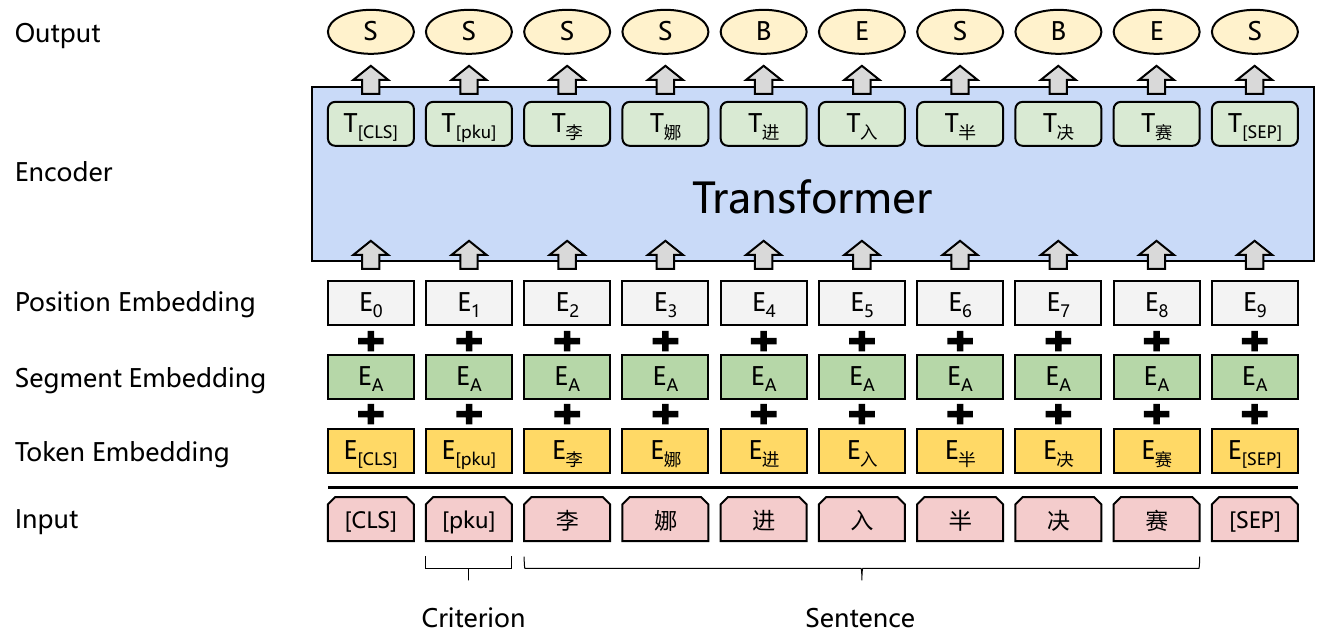}
    \caption{The unified framework of our proposed model, with shared encoder and decoder for different criteria. The input is composed of criterion and sentence, where the criterion can vary with the same sentence. The output is a corresponding sequence of segmentation labels of given criterion.}
    \label{fig:unified}
\end{figure*}

To better leverage the common segmentation knowledge shared by multiple criteria, \textsc{Meta}\textsc{Seg} employs a unified architecture based on the widely-used Transformer network~\cite{vaswani-2017-attention} with shared encoder and decoder for all different criteria, as illustrated in Figure~\ref{fig:unified}.

The input for the unified architecture is an augmented sentence, which is composed of a specific criterion token plus the original sentence to represent both criterion and text information.
In embedding layer, the augmented sentence is transformed into input representations by summing the token, segment and position embeddings.
The Transformer network is used as the shared encoder layer, encoding the input representations into hidden representations through blocks of multi-head attention and position-wise feed-forward modules~\cite{vaswani-2017-attention}.
Then a shared linear decoder with softmax is followed to map hidden representations to the probability distribution of segmentation labels.
The segmentation labels consist of four CWS labels $\{B,M,E,S\}$, denoting the beginning, middle, ending and single of a word respectively.

Formally, the unified architecture can be concluded as a probabilistic model $P_{\theta} (Y|X)$, which represents the probability of the segmentation label sequence $Y$ given the augmented input sentence $X$.
The model parameters $\theta$ are invariant of any criterion $c$, and would capture the common segmentation knowledge shared by different criteria.

\subsection{Multi-Criteria Pre-training with Meta Learning}
\label{subsec:multi}

In this part, we describe the multi-criteria pre-training task with meta learning algorithm.
We construct a multi-criteria pre-training task, to fully mine the shared prior segmentation knowledge of different criteria.
Meanwhile, to alleviate the discrepancy between pre-trained models and downstream CWS tasks, meta learning algorithm~\cite{finn-2017-maml} is used for pre-training optimization of \textsc{Meta}\textsc{Seg}.

\paragraph{Multi-Criteria Pre-training Task}
As mentioned in Section~\ref{sec:intro}, there are already a variety of existing CWS corpora~\cite{emerson-2005-second, jin-2008-fourth}.
These CWS corpora usually have inconsistent segmentation criteria, where human-annotated data is insufficient for each criterion.
Each criterion is used to fine-tune a CWS model separately on a relatively small dataset and ignores the shared knowledge of different criteria.
But in our multi-criteria pre-training task, multiple criteria are jointly used for pre-training to capture the common segmentation knowledge shared by different existing criteria.

First, nine public CWS corpora (see Section ~\ref{subsec:dataset}) of diverse segmentation criteria are merged as a joint multi-criteria pre-training corpus $D_T$.
Every sentence under each criterion is augmented with the corresponding criterion, and then incorporated into the joint multi-criteria pre-training corpus.
To represent criterion information, we add a specific criterion token in front of the input sentence, such as \verb|[pku]| for PKU criterion~\cite{emerson-2005-second}.
We also add \verb|[CLS]| and \verb|[SEP]| token to sentence beginning and ending respectively like ~\newcite{devlin-2019-bert}.
This augmented input sentence represents both criterion and text information, as shown in Figure~\ref{fig:unified}.

Then, we randomly pick 10\% sentences from the joint multi-criteria pre-training corpus $D_T$ and replace their criterion tokens with a special token \verb|[unc]|, which means undefined criterion.
With this design, the undefined criterion token \verb|[unc]| would learn criterion-independent segmentation knowledge and help to transfer such knowledge to downstream CWS tasks.

Finally, given a pair of augmented sentence $X$ and segmentation labels $Y$ from the joint multi-criteria pre-training corpus $D_T$, our unified architecture (Section~\ref{subsec:unified}) predicts the the probability of segmentation labels $P_{\theta} (Y|X)$.
We use the normal negative log-likelihood (NLL) loss as objective function for this multi-criteria pre-training task:
\begin{equation}
    L(\theta; D_T) = - \sum_{X,Y \in D_T} \log P_{\theta} (Y|X)
\end{equation}

\paragraph{Meta Learning Algorithm}

\begin{figure}[htbp]
    \centering
    \subfigure[Pre-training without meta learning]{
    \includegraphics[width=0.4\columnwidth]{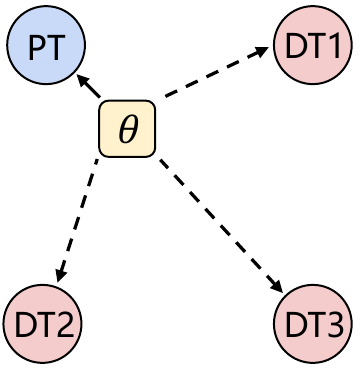}
    }
    \quad
    \subfigure[Pre-training with meta learning]{
    \includegraphics[width=0.4\columnwidth]{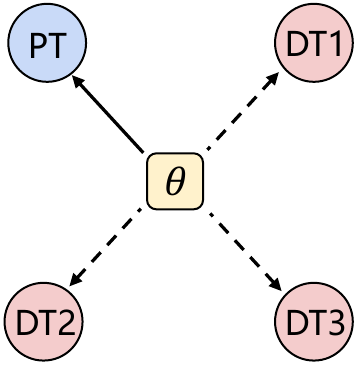}
    }
    \caption{Pre-training with and without meta learning. PT represents the multi-criteria pre-training task, while solid line represents the pre-training phase. DT represents the downstream CWS task, while dashed line represents the fine-tuning phase. $\theta$ represents pre-trained model parameters.}
    \label{fig:meta}
\end{figure}

The objective of most PTMs is to maximize its performance on pre-training tasks~\cite{devlin-2019-bert}, which would lead to the discrepancy between pre-trained models and downstream tasks.
Besides, pre-trained CWS model from multi-criteria pre-training task could still have discrepancy with downstream unseen criteria, because downstream criteria may not exist in pre-training.
To alleviate the above discrepancy, we utilize meta learning algorithm~\cite{lv-2020-pretraining} for pre-training optimization of \textsc{Meta}\textsc{Seg}.
The main objective of meta learning is to maximize generalization performance on potential downstream tasks, which prevents pre-trained models from overfitting on pre-training tasks.
As shown in Figure~\ref{fig:meta}, by introducing meta learning algorithm, pre-trained models would have less discrepancy with downstream tasks instead of inclining towards pre-training tasks.

The meta learning algorithm treats pre-training task $T$ as one of the downstream tasks.
It tries to optimize meta parameters $\theta_0$, from which we can get the task-specific model parameters $\theta_k$ by $k$ gradient descent steps over the training data $D_T^{train}$ on task $T$,
\begin{equation}
\begin{aligned}
    \theta_1 & = \theta_{0} - \alpha \nabla_{\theta_{0}} L_{T} ( \theta_{0}; D_{T,1}^{train} ) , \\
    ..., \\
    \theta_k & = \theta_{k-1} - \alpha \nabla_{\theta_{k-1}} L_{T} ( \theta_{k-1}; D_{T,k}^{train} ) ,
\end{aligned}
\label{eq:meta-train}
\end{equation}
where $\alpha$ is learning rate, $D_{T,i}^{train}$ is the $i$-th batch of training data.
Formally, task-specific parameters $\theta_k$ can be denoted as a function of meta parameters $\theta_0$ as follows: $\theta_k = f_k ( \theta_0 )$.

To maximize the generalization performance on task $T$, we should optimize meta parameters $\theta_0$ on the batch of test data $D_T^{test}$,
\begin{equation}
\begin{aligned}
    \theta_0^* &= \mathop{\arg\min}_{\theta_0} L_{T} (\theta_{k}; D_T^{test} ) \\
    &= \mathop{\arg\min}_{\theta_0} L_{T} ( f_k ( \theta_0 ); D_T^{test} ) .
\end{aligned}
\end{equation}

The above meta optimization could be achieved by gradient descent, so the update rule for meta parameters $\theta_0$ is as follows:
\begin{equation}
  \theta_0' = \theta_0 - \beta \nabla_{\theta_0} L_{T} ( \theta_k ; D_T^{test} ) , \\
\label{eq:meta-update}
\end{equation}
where $\beta$ is the meta learning rate.
The gradient in Equation~\ref{eq:meta-update} can be rewritten as:
\begin{equation} \small
\begin{aligned}
  & \nabla_{\theta_0} L_{T} ( \theta_k ; D_T^{test} ) \\
  & = \nabla_{\theta_k} L_{T} ( \theta_k ; D_T^{test} ) \times \nabla_{\theta_{k-1}} \theta_k \times \cdots \nabla_{\theta_0} \theta_1 \\
  & = \nabla_{\theta_k} L_{T} ( \theta_k ; D_T^{test} ) \prod_{j=1}^k ( I - \alpha \nabla_{\theta_{j-1}}^2 L_{T} (\theta_{j-1}; D_{T,j}^{train})  ) \\
  & \approx \nabla_{\theta_k} L_{T} ( \theta_k ; D_T^{test} ) ,
\end{aligned}
\label{eq:meta-test}
\end{equation}
where the last step in Equation~\ref{eq:meta-test} adopts first-order approximation for computational simplification~\cite{finn-2017-maml}.

Specifically, the meta learning algorithm for pre-training optimization is described in Algorithm~\ref{alg:meta}.
It can be divided into two stages:
i) meta train stage, which updates task-specific parameters by $k$ gradient descent steps over training data;
ii) meta test stage, which updates meta parameters by one gradient descent step over test data.
Hyper-parameter $k$ is the number of gradient descent steps in meta train stage.
The meta learning algorithm degrades to normal gradient descent algorithm when $k=0$.
The returned meta parameters $\theta_0$ are used as the pre-trained model parameters for \textsc{Meta}\textsc{Seg}.

\begin{algorithm}[htbp]
\centering
\caption{Meta Learning for Pre-training Optimization}
\label{alg:meta}
\begin{algorithmic}[1]
    \REQUIRE Distribution over pre-training task $p(T)$, initial meta parameters $\theta_0$, objective function $L$
    \REQUIRE Learning rate $\alpha$, meta learning rate $\beta$, meta train steps $k$
    \FOR{$epoch = 1,2, ...$}
        \STATE Sample k training data batches $D_T^{train}$ from $p(T)$
        \FOR{$j = 1,2, ...,k$}
            \STATE $\theta_j \leftarrow \theta_{j-1} - \alpha \nabla_{\theta_{j-1}} L_{T} ( \theta_{j-1}; D_{T,j}^{train} )$
        \ENDFOR
        \STATE Sample test data batch $D_T^{test}$ from $p(T)$
        \STATE $\theta_0 \leftarrow \theta_0 - \beta \nabla_{\theta_k} L_{T} ( \theta_k ; D_T^{test} )$
    \ENDFOR
    \RETURN Meta parameters $\theta_0$
\end{algorithmic}
\end{algorithm}

\subsection{Downstream Fine-tuning}
\label{subsec:fine}

After pre-training phase mentioned in Section~\ref{subsec:multi}, we obtain the pre-trained model parameters $\theta_0$, which capture prior segmentation knowledge and have less discrepancy with downstream CWS tasks.
We fine-tune these pre-trained parameters $\theta_0$ on downstream CWS corpus, to transfer the prior segmentation knowledge.

For format consistency, we process the sentence from the given downstream corpus in the same way as Section~\ref{subsec:multi}, by adding the criterion token \verb|[unc]|, beginning token \verb|[CLS]| and ending token beginning token \verb|[SEP]|.
The undefined criterion token \verb|[unc]| is used in fine-tuning phase instead of the downstream criterion itself, because the downstream criterion usually doesn't exist in pre-training phase and the pre-trained model has no information about it.

\section{Experiment}
\label{sec:exp}

\subsection{Experimental Settings}
\label{subsec:dataset}

\paragraph{Datasets}
We collect twelve publicly available CWS datasets, with each dataset representing a unique segmentation criterion.
Among all datasets, we have PKU, MSRA, CITYU, AS from SIGHAN2005~\cite{emerson-2005-second}, CKIP, NCC, SXU from SIGHAN2008~\cite{jin-2008-fourth}, CTB6 from~\newcite{xue-2005-ctb}, WTB from~\newcite{wang-2014-dependency}, UD from~\newcite{zeman-2017-conll}, ZX from~\newcite{zhang-2014-type} and CNC~\footnote{\url{http://corpus.zhonghuayuwen.org/}}.

WTB, UD, ZX datasets are kept for downstream fine-tuning phase, while the other nine datasets are combined into the joint multi-criteria pre-training corpus (Section~\ref{subsec:multi}), which amounts to nearly 18M words.

For CTB6, WTB, UD, ZX and CNC datasets, we use the official data split of training, development, and test sets.
For the rest, we use the official test set and randomly pick 10\% samples from the training data as the development set.
We pre-process all these datasets following four procedures: 
\begin{enumerate}
    \item Convert traditional Chinese datasets into simplified, such as CITYU, AS and CKIP;
\item Convert full-width tokens into half-width;
\item Replace continuous English letters and digits with unique tokens;
\item Split sentences into shorter clauses by punctuation.
\end{enumerate}
Table~\ref{tbl:dataset} presents the statistics of processed datasets.

\begin{table*}[t] \small
\centering
\begin{tabular}{lccccc}
\toprule
\textbf{Corpus} & \textbf{\#Train Words} & \textbf{\#Dev Words} & \textbf{\#Test Words} & \textbf{OOV Rate} & \textbf{Avg. Length} \\ \toprule
PKU & 999,823 & 110,124 & 104,372 & 3.30\% & 10.6 \\
MSRA & 2,133,674 & 234,717 & 106,873 & 2.11\% & 11.3 \\
CITYU & 1,308,774 & 146,856 & 40,936 & 6.36\% & 11.0 \\
AS & 4,902,887 & 546,694 & 122,610 & 3.75\% & 9.7 \\
CKIP & 649,215 & 72,334 & 90,678 & 7.12\% & 10.5 \\
NCC & 823,948 & 89,898 & 152,367 & 4.82\% & 10.0 \\
SXU & 475,489 & 52,749 & 113,527 & 4.81\% & 11.1 \\
CTB6 & 678,811 & 51,229 & 52,861 & 5.17\% & 12.5 \\
CNC & 5,841,239 & 727,765 & 726,029 & 0.75\% & 9.8 \\ \midrule
WTB & 14,774 & 1,843 & 1,860 & 15.05\% & 28.2 \\
UD & 98,607 & 12,663 & 12,012 & 11.04\% & 11.4 \\
ZX & 67,648 & 20,393 & 67,648 & 6.48\% & 8.2 \\ \bottomrule
\end{tabular}
\caption{Statistics of datasets. The first block corresponds to the pre-training criteria. The second block corresponds to downstream criteria, which are invisible in pre-training phase.}
\label{tbl:dataset}
\end{table*}

\paragraph{Hyper-Parameters}
We employ \textsc{Meta}\textsc{Seg} with the same architecture as BERT-Base~\cite{devlin-2019-bert}, which has 12 transformer layers, 768 hidden sizes and 12 attention heads.

In pre-training phase, \textsc{Meta}\textsc{Seg} is initialized with released parameters of Chinese BERT-Base model~\footnote{\url{https://github.com/google-research/bert}} and then pre-trained with the multi-criteria pre-training task. 
Maximum input length is 64, with batch size 64, and dropout rate 0.1.
We adopt AdamW optimizer~\cite{loshchilov-2018-decoupled} with $\beta_1 = 0.9, \beta_2 = 0.999$ and weight decay rate of 0.01.
The optimizer is implemented by meta learning algorithm, where both learning rate $\alpha$ and meta learning rate $\beta$ are set to 2e-5 with a linear warm-up proportion of 0.1.
The meta train steps are selected to $k=1$ according to downstream performance.
Pre-training process runs for nearly 127,000 meta test steps, amounting to $(k+1) * 127,000$ gradient descent steps, which takes about 21 hours on one NVIDIA Tesla V100 32GB GPU card.

In fine-tuning phase, we set maximum input length to 64 for all criteria but 128 for WTB, with batch size 64.
We fine-tune \textsc{Meta}\textsc{Seg} with AdamW optimizer of the same settings as pre-training phase without meta learning.
\textsc{Meta}\textsc{Seg} is fine-tuned for 5 epochs on each downstream dataset.

In low-resource settings, experiments are performed on WTB dataset, with maximum input length 128.
We evaluate \textsc{Meta}\textsc{Seg} at sampling rates of 1\%, 5\%, 10\%, 20\%, 50\%, 80\%.
Batch size is 1 for 1\% sampling and 8 for the rest.
We keep other hyper-parameters the same as those of fine-tuning phase.

The standard average F1 score is used to evaluate the performance of all models.

\subsection{Results on Pre-training Criteria}
\label{subsec:pretrain}

After pre-training, we fine-tune \textsc{Meta}\textsc{Seg} on each pre-training criterion.
Table~\ref{tbl:exp-pretrain} shows F1 scores on test sets of nine pre-training criteria in two blocks. 
The first block displays the performance of previous works.
The second block displays three models implemented by us: \textbf{BERT-Base} is the fine-tuned model initialized with official BERT-Base parameters.
\textbf{\textsc{Meta}\textsc{Seg} (w/o fine-tune)} is our proposed pre-trained model directly used for inference without fine-tuning.
\textbf{\textsc{Meta}\textsc{Seg}} is the fine-tuned model initialized with pre-trained \textsc{Meta}\textsc{Seg} parameters.

\begin{table*}[thbp] \small
\centering
\begin{tabular}{lcccccccccc}
\toprule
\textbf{Models} & \textbf{PKU} & \textbf{MSRA} & \textbf{CITYU} & \textbf{AS} & \textbf{CKIP} & \textbf{NCC} & \textbf{SXU} & \textbf{CTB6} & \textbf{CNC} & \textbf{Avg.} \\ \toprule
\newcite{chen-2017-adversarial} & 94.32 & 96.04 & 95.55 & 94.64 & 94.26 & 92.83 & 96.04 & - & - & - \\
\newcite{ma-2018-state} & 96.10 & 97.40 & 97.20 & 96.20 & - & - & - & 96.70 & - & - \\
\newcite{he-2019-effective} & 95.78 & 97.35 & 95.60 & 95.47 & 95.73 & 94.34 & 96.49 & - & - & - \\
\newcite{gong-2019-switch} & 96.15 & 97.78 & 96.22 & 95.22 & 94.99 & 94.12 & 97.25 & - & - & - \\
\newcite{yang-2019-subword} & 95.80 & 97.80 & - & - & - & - & - & 96.10 & - & - \\
\newcite{meng-2019-glyce} & 96.70 & 98.30 & 97.90 & 96.70 & - & - & - & - & - & - \\
\newcite{yang-2019-bert} & 96.50 & 98.40 & - & - & - & - & - & - & - & - \\
\newcite{duan-2020-attention} & 95.50 & 97.70 & 96.40 & 95.70 & - & - & - & - & - & - \\
\newcite{huang-2019-toward} & \textbf{97.30} & 98.50 & 97.80 & 97.00 & - & - & 97.50 & 97.80 & 97.30 & - \\
\newcite{qiu-2019-multi} & 96.41 & 98.05 & 96.91 & 96.44 & 96.51 & 96.04 & 97.61 & - & - & - \\ 
\newcite{tian-2020-improving} & 96.53 & 98.40 & 97.93 & 96.62 & - & - & - & 97.25 & - & - \\ 
\midrule
BERT-Base (ours) & 96.72 & 98.25 & 98.19 & 96.93 & 96.49 & 96.13 & 97.61 & 97.85 & 97.45 & 97.29 \\
\textsc{Meta}\textsc{Seg} (w/o fine-tune) & 96.76 & 98.02 & 98.12 & \textbf{97.04} & \textbf{96.81} & 97.21 & 97.51 & 97.87 & 97.25 & 97.40 \\
\textsc{Meta}\textsc{Seg} & 96.92 & \textbf{98.50} & \textbf{98.20} & 97.01 & 96.72 & \textbf{97.24} & \textbf{97.88} & \textbf{97.89} & \textbf{97.55} & \textbf{97.55} \\ \bottomrule
\end{tabular}
\caption{F1 scores on test sets of pre-training criteria. The first block displays results from previous works. The second block displays three models implemented by us.}
\label{tbl:exp-pretrain}
\end{table*}

From the second block, we observe that fine-tuned \textsc{Meta}\textsc{Seg} could outperform fine-tuned BERT-Base on each criterion, with 0.26\% improvement on average.
It shows that \textsc{Meta}\textsc{Seg} is more effective when fine-tuned for CWS.
Even without fine-tuning, \textsc{Meta}\textsc{Seg} (w/o fine-tune) still behaves better than fine-tuned BERT-Base model, indicating that our proposed pre-training approach is the key factor for the effectiveness of \textsc{Meta}\textsc{Seg}.
Fine-tuned \textsc{Meta}\textsc{Seg} performs better than that of no fine-tuning, showing that downstream fine-tuning is still necessary for the specific criterion.
Furthermore, \textsc{Meta}\textsc{Seg} can achieve state-of-the-art results on eight of nine pre-training criteria, demonstrating the effectiveness of our proposed methods.

\subsection{Results on Downstream Criteria}
\label{subsec:unseen}

To evaluate the knowledge transfer ability of \textsc{Meta}\textsc{Seg}, we preform experiments on three unseen downstream criteria which are absent in pre-training phase.
Table~\ref{tbl:exp-unseen} shows F1 scores on test sets of three downstream criteria.
The first block displays previous works on these downstream criteria, while the second block displays three models implemented by us (see Section~\ref{subsec:pretrain} for details).

\begin{table}[htbp] \small
\centering
\begin{tabular}{lcccc}
\toprule
\textbf{Models} & \textbf{WTB} & \textbf{UD} & \textbf{ZX} & \textbf{Avg.} \\ \toprule
\newcite{ma-2018-state} & - & 96.90 & - & - \\
\newcite{huang-2019-toward} & 93.20 & 97.80 & 97.10 & 96.03 \\ \midrule
BERT-Base (ours) & 93.00 & 98.32 & 97.06 & 96.13 \\
\begin{tabular}[l]{@{}l@{}}\textsc{Meta}\textsc{Seg}\\ (w/o fine-tune)\end{tabular} & 89.53 & 83.84 & 88.48 & 87.28 \\
\textsc{Meta}\textsc{Seg} & \textbf{93.97} & \textbf{98.49} & \textbf{97.22} & \textbf{96.56} \\ \bottomrule
\end{tabular}
\caption{F1 scores on test sets of downstream criteria.}
\label{tbl:exp-unseen}
\end{table}

Results show that \textsc{Meta}\textsc{Seg} outperforms the previous best model by 0.53\% on average, achieving new state-of-the-art performance on three downstream criteria.
Moreover, \textsc{Meta}\textsc{Seg} (w/o fine-tune) actually preforms zero-shot inference on downstream criteria and still achieves 87.28\% average F1 score.
This shows that \textsc{Meta}\textsc{Seg} does learn some common prior segmentation knowledge in pre-training phase, even if it doesn't see these downstream criteria before.

Compared with BERT-Base, \textsc{Meta}\textsc{Seg} has the same architecture but different pre-training tasks.
It can be easily observed that \textsc{Meta}\textsc{Seg} with fine-tuning outperforms BERT-Base by 0.43\% on average.
This indicates that \textsc{Meta}\textsc{Seg} could indeed alleviate the discrepancy between pre-trained models and downstream CWS tasks than BERT-Base.

\subsection{Ablation Studies}
\label{subsec:ablation}

We perform further ablation studies on the effects of meta learning (ML) and multi-criteria pre-training (MP), by removing them consecutively from the complete \textsc{Meta}\textsc{Seg} model.
After removing both of them, \textsc{Meta}\textsc{Seg} degrades into the normal BERT-Base model.
F1 scores for ablation studies on three downstream criteria are illustrated in Table~\ref{tbl:ablation}.

We observe that the average F1 score drops by 0.09\% when removing the meta learning algorithm (-ML), and continues to drop by 0.34\% when removing the multi-criteria pre-training task (-ML-MP).
It demonstrates that meta learning and multi-criteria pre-training are both significant for the effectiveness of \textsc{Meta}\textsc{Seg}.

\begin{table}[htbp] \small
\centering
\begin{tabular}{lcccc}
\toprule
\textbf{Models} & \textbf{WTB} & \textbf{UD} & \textbf{ZX} & \textbf{Avg.} \\ \toprule
\textsc{Meta}\textsc{Seg} & \textbf{93.97} & \textbf{98.49} & \textbf{97.22} & \textbf{96.56} \\
-ML & 93.71 & 98.49 & 97.22 & 96.47 \\
-ML-MP & 93.00 & 98.32 & 97.06 & 96.13 \\ \bottomrule
\end{tabular}
\caption{F1 scores for ablation studies on downstream criteria. -ML indicates \textsc{Meta}\textsc{Seg} without meta learning. -ML-MP indicates \textsc{Meta}\textsc{Seg} without meta learning and multi-criteria pre-training.}
\label{tbl:ablation}
\end{table}

\subsection{Low-Resource Settings}
\label{subsec:low}

\begin{table*}[thbp] \small
\centering
\begin{tabular}{lccccccc}
\toprule
Sampling Rates & 1\% & 5\% & 10\% & 20\% & 50\% & 80\% & 100\% \\ \midrule
\#Instances & 8 & 40 & 81 & 162 & 406 & 650 & 813 \\ \midrule
BERT-Base (ours) & 85.40 & 87.83 & 90.46 & 91.15 & 92.80 & 93.14 & 93.00 \\
\textsc{Meta}\textsc{Seg} & \textbf{91.60} & \textbf{92.29} & \textbf{92.54} & \textbf{92.63} & \textbf{93.45} & \textbf{94.11} & \textbf{93.97} \\ \bottomrule
\end{tabular}
\caption{F1 scores on WTB test set in low-resource settings.}
\label{tbl:exp-low-resource}
\end{table*}

To better explore the downstream generalization ability of \textsc{Meta}\textsc{Seg}, we perform experiments on the downstream WTB criterion in low-resource settings.
Specifically, we randomly sample a given rate of instances from the training set and fine-tune the pre-trained \textsc{Meta}\textsc{Seg} model on down-sampling training sets.
These settings imitate the realistic low-resource circumstance where human-annotated data is insufficient.

The performance at different sampling rates is evaluated on the same WTB test set and reported in Table~\ref{tbl:exp-low-resource}.
Results show that \textsc{Meta}\textsc{Seg} outperforms BERT-Base at every sampling rate.
The margin is larger when the sampling rate is lower, and reaches 6.20\% on 1\% sampling rate.
This demonstrates that \textsc{Meta}\textsc{Seg} could generalize better on the downstream criterion in low-resource settings.

When the sampling rate drops from 100\% to 1\%, F1 score of BERT-Base decreases by 7.60\% while that of \textsc{Meta}\textsc{Seg} only decreases by 2.37\%.
The performance of \textsc{Meta}\textsc{Seg} at 1\% sampling rate still reaches 91.60\% with only 8 instances, comparable with performance of BERT-Base at 20\% sampling rate.
This indicates that \textsc{Meta}\textsc{Seg} can make better use of prior segmentation knowledge and learn from less amount of data.
It shows that \textsc{Meta}\textsc{Seg} would reduce the need of human annotation significantly.

\subsection{Out-of-Vocabulary Recalls}
\label{subsec:oov}

Out-of-Vocabulary (OOV) words denote the words which exist in inference phase but don't exist in training phase.
OOV words are a critical cause of errors on CWS tasks.
We evaluate recalls of OOV words on test sets of all twelve criteria in Table~\ref{tbl:exp-oov}.

\begin{table*}[thbp] \small
\centering
\setlength{\tabcolsep}{1.5mm}{
\begin{tabular}{lccccccccccccc}
\toprule
\textbf{Models} & \textbf{PKU} & \textbf{MSRA} & \textbf{CITYU} & \textbf{AS} & \textbf{CKIP} & \textbf{NCC} & \textbf{SXU} & \textbf{CTB6} & \textbf{CNC} & \textbf{WTB} & \textbf{UD} & \textbf{ZX} & \textbf{Avg.} \\ \toprule
BERT-Base & 80.15 & 81.03 & 90.62 & 79.60 & \textbf{84.48} & 79.64 & 84.75 & 89.10 & 61.18 & 83.57 & 93.36 & \textbf{87.69} & 82.93 \\
\textsc{Meta}\textsc{Seg} & \textbf{80.90} & \textbf{83.03} & \textbf{90.66} & \textbf{80.89} & 84.42 & \textbf{84.14} & \textbf{85.98} & \textbf{89.21} & \textbf{61.90} & \textbf{85.00} & \textbf{93.59} & 87.33 & \textbf{83.92} \\ \bottomrule
\end{tabular}}
\caption{OOV recalls on test sets of all criteria.}
\label{tbl:exp-oov}
\end{table*}

Results show that \textsc{Meta}\textsc{Seg} outperforms BERT-Base on ten of twelve criteria and improves OOV recall by 0.99\% on average.
This indicates that \textsc{Meta}\textsc{Seg} could benefit from our proposed pre-training methodology and recognize more OOV words in inference phase.

\subsection{Visualization}
\label{subsec:visual}

To visualize the discrepancy between pre-trained models and downstream criteria, we plot similarities of three downstream criteria with \textsc{Meta}\textsc{Seg} and BERT.

\begin{figure}[htbp]
    \centering
    \includegraphics[width=0.75\columnwidth]{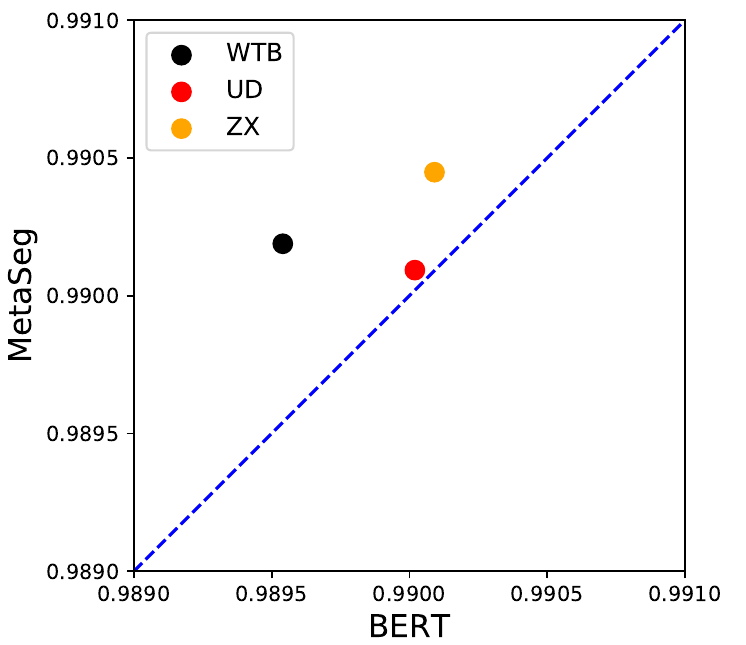}
    \caption{Cosine similarities between three downstream criteria and two pre-trained models. The dashed line indicates the positions where one criterion has equal similarities with two pre-trained models.}
    \label{fig:visual}
\end{figure}

Specifically, we extract the criterion token embeddings of three downstream criteria WTB, UD and ZX.
We also extract the undefined criterion token embeddings of \textsc{Meta}\textsc{Seg} and BERT as representations of these two pre-trained models.
We compute cosine similarities between three criteria embeddings and two pre-trained model embeddings, and illustrate them in Figure~\ref{fig:visual}.

We can observe that similarities of all three downstream criteria lie above the dashed line, indicating that all three downstream criteria are more similar to \textsc{Meta}\textsc{Seg} than BERT.
The closer one criterion is to the upper left corner, the more similar it is to \textsc{Meta}\textsc{Seg}.
Therefore, we can conclude that WTB is the most similar criterion to \textsc{Meta}\textsc{Seg} among all these criteria, which qualitatively corresponds to the phenomenon that WTB criterion has the largest performance boost in Table~\ref{tbl:exp-unseen}.
The above visualization results show that our proposed approach could solidly alleviate the discrepancy between pre-trained models and downstream CWS tasks.
Thus \textsc{Meta}\textsc{Seg} is more similar to downstream criteria.

\section{Conclusion}
\label{sec:conclusion}

In this paper, we propose a CWS-specific pre-trained model \textsc{Meta}\textsc{Seg}, which employs a unified architecture and incorporates meta learning algorithm into a multi-criteria pre-training task.
Experiments show that \textsc{Meta}\textsc{Seg} could make good use of common prior segmentation knowledge from different existing criteria, and alleviate the discrepancy between pre-trained models and downstream CWS tasks.
\textsc{Meta}\textsc{Seg} also gives better generalization ability in low-resource settings, and achieves new state-of-the-art performance on twelve CWS datasets.

In the future, we will explore unsupervised pre-training methods for CWS, and whether representations from the pre-trained CWS model are helpful for other related NLP tasks.



\bibliography{anthology,custom}
\bibliographystyle{acl_natbib}




\end{CJK}
\end{document}